\documentclass[sigconf, language=french,
language=german, language=spanish, language=english]{acmart}

\usepackage{subfigure}
\usepackage{caption}
\usepackage{makecell}
\usepackage{float}

\AtBeginDocument{%
  }

\setcopyright{acmcopyright}
\copyrightyear{2018}
\acmYear{2018}
\acmDOI{XXXXXXX.XXXXXXX}

\acmConference[Conference acronym 'XX]{Make sure to enter the correct
  conference title from your rights confirmation emai}{June 03--05,
  2018}{Woodstock, NY}
\acmPrice{15.00}
\acmISBN{978-1-4503-XXXX-X/18/06}

\begin{document}

\title{Perturbation-Based Two-Stage Multi-Domain Active Learning}


% \author{Anonymous Authors}

\author{Rui He}
\affiliation{%
  \institution{University of Birmingham}
  % \streetaddress{1 Th{\o}rv{\"a}ld Circle}
  \city{Birmingham}
  \country{United Kingdom}}
\affiliation{%
  \institution{Southern University of Science and Technology}
  % \streetaddress{1 Th{\o}rv{\"a}ld Circle}
  \city{Shenzhen}
  \country{China}}
\email{rxh877@bham.ac.uk}

\author{Zeyu Dai}
\affiliation{%
  \institution{The Hong Kong Polytechnic University}
  % \streetaddress{1 Th{\o}rv{\"a}ld Circle}
  \city{Hong Kong}
  \country{China}
}
\affiliation{%
  \institution{Southern University of Science and Technology}
  % \streetaddress{1 Th{\o}rv{\"a}ld Circle}
  \city{Shenzhen}
  \country{China}}
\email{ze-yu.dai@connect.polyu.hk}

\author{Shan He}
\affiliation{%
  \institution{University of Birmingham}
  % \streetaddress{1 Th{\o}rv{\"a}ld Circle}
  \city{Birmingham}
  \country{United Kingdom}}
\email{s.he@cs.bham.ac.uk}

\author{Ke Tang}
\affiliation{%
  \institution{Southern University of Science and Technology}
  % \streetaddress{1 Th{\o}rv{\"a}ld Circle}
  \city{Shenzhen}
  \country{China}}
\email{tangk3@sustech.edu.cn}

\renewcommand{\shortauthors}{Anonymous Authors}

\begin{abstract}
  In multi-domain learning (MDL) scenarios, high labeling effort is required due to the complexity of collecting data from various domains.
  Active Learning (AL) presents an encouraging solution to this issue by annotating a smaller number of highly informative instances, thereby reducing the labeling effort.
  Previous research has relied on conventional AL strategies for MDL scenarios, which underutilize the domain-shared information of each instance during the selection procedure.
  To mitigate this issue, we propose a novel perturbation-based two-stage multi-domain active learning (P2S-MDAL) method incorporated into the well-regarded ASP-MTL model.
  Specifically, P2S-MDAL involves allocating budgets for domains and establishing regions for diversity selection, which are further used to select the most cross-domain influential samples in each region.
  A perturbation metric has been introduced to evaluate the robustness of the shared feature extractor of the model, facilitating the identification of potentially cross-domain influential samples.
  Experiments are conducted on three real-world datasets, encompassing both texts and images.
  The superior performance over conventional AL strategies shows the effectiveness of the proposed strategy.
  Additionally, an ablation study has been carried out to demonstrate the validity of each component.
  Finally, we outline several intriguing potential directions for future MDAL research, thus catalyzing the field's advancement.
\end{abstract}

\begin{CCSXML}
<ccs2012>
   <concept>
       <concept_id>10010147.10010257.10010282.10011304</concept_id>
       <concept_desc>Computing methodologies~Active learning settings</concept_desc>
       <concept_significance>500</concept_significance>
       </concept>
   <concept>
       <concept_id>10010147.10010257.10010258.10010262.10010277</concept_id>
       <concept_desc>Computing methodologies~Transfer learning</concept_desc>
       <concept_significance>500</concept_significance>
       </concept>
   <concept>
       <concept_id>10010147.10010257.10010258.10010262</concept_id>
       <concept_desc>Computing methodologies~Multi-task learning</concept_desc>
       <concept_significance>300</concept_significance>
       </concept>
 </ccs2012>
\end{CCSXML}

\ccsdesc[500]{Computing methodologies~Active learning settings}
\ccsdesc[500]{Computing methodologies~Transfer learning}
\ccsdesc[300]{Computing methodologies~Multi-task learning}

\keywords{active learning, multi-domain learning}



\maketitle

\section{Introduction}

% Introduction:
In practical applications, aggregating data from diverse sources is a common practice for accomplishing specific tasks.
These data sources, often called "domains," exhibit distinct distributions.
For instance, sentiment analysis may involve data collection from various social media platforms such as Twitter, Facebook, and Weibo.
In image classification, different styles of images \cite{PACs}, including sketches, cartoons, art paintings, and camera photos, represent distinct domains.
Each domain possesses unique characteristics and contexts while containing sharable intertwined information.
% Although building separate models for each domain is feasible, it fails to exploit the potential benefits of shared information.
If appropriately harnessed, this information could significantly improve the performance of machine learning models.
Multi-domain learning (MDL) \cite{MDL} aims to simultaneously learn across various domains, leveraging shared knowledge to enhance overall performance.
Empirically, MDL outperforms both single-domain learning and joint learning across multiple domains \cite{mdal} in many real-world applications.
% The Issue of High Labeling Effort in Multi-Domain Learning:
However, high labeling effort represents a challenge in MDL, since data need to be collected from multiple domain experts.
% For example, in the context of multi-domain medical image classification, the manual annotation cost associated with different field experts is just one of the challenges \cite{mdl-medical-image}.
Additionally, varying legal, privacy, and ethical requirements and annotation tools across domains make the multi-domain data collection process more arduous.
Therefore, \textbf{it is crucial to minimize the labeling effort in MDL.}

% Active Learning as a Solution:
Active learning (AL) \cite{AL-Comparative-Survey} presents a promising solution for reducing the labeling effort in machine learning tasks.
By iteratively selecting informative samples for annotation, AL achieves comparable performance to random selection while requiring significantly less labeling effort.
Several studies have explored the application of AL to reduce labeling effort in MDL, known as multi-domain active learning (\textbf{MDAL}) \cite{MDAL-text}.
Most works simply adapt conventional single-domain AL strategies to MDL models \cite{mdal,MDAL-text}, which leads to noticeable improvements.
They mix all the evaluations from different domains and select the ones with the highest evaluation score.
Nonetheless, solely applying single-domain AL to MDL models is suboptimal, as the domain-shared information remains underutilized in item selection.
Besides, the mixed scores from different domains are incomparable, potentially leading to biased selection.
To the best of our knowledge, no existing work has designed AL strategies specifically for MDL to tackle these issues.
Thus, a natural question arises: \textbf{how can we design effective AL strategies tailored explicitly for MDL?}

% Research Objective and Contributions:
This paper proposes a novel perturbation-based two-stage multi-domain active learning (P2S-MDAL) strategy, which builds upon the classical and renowned MDL model, ASP-MTL \cite{ASP-MTL}.
In the first stage, we allocate a budget to each domain to ensure a fair in-domain comparison and establish regions for diversity selection.
Subsequently, within each region, we further select the most cross-domain influential samples for annotation.
The influence evaluation is based on perturbations, a novel metric that assesses the robustness of the shared feature extractor of ASP-MTL.
The underlying intuition is that if a sample is more informative to the shared extractor, it will be more vulnerable to perturbations, i.e., the perturbed sample will likely have more distinct outputs compared to the original one.
Consequently, such examples, which are less learned by the shared feature extractor, could be more influential to all the domains.
Experimental results on three real-world datasets, encompassing texts and images, demonstrate the superiority of the proposed P2S-MDAL strategy over conventional AL strategies.

The main contributions of this paper are summarized as follows:

\begin{itemize}
  \item We introduce a novel AL strategy called P2S-MDAL, the first strategy tailored for the MDL scenario upon the renowned ASP-MTL model. Experimental results indicate noticeable improvements over conventional AL strategies.
  \item We employ perturbations to evaluate the cross-domain influences of instances in AL. This perspective offers a fresh viewpoint in assessing the potential of individual instances.
  \item We highlight several intriguing research directions for MDAL.
\end{itemize}

% However, these works are limited to the homogeneous MDAL setting, where the domains share the same feature space and label space.

\section{Problem Formulation}

Same to AL, MDAL could be formulated as a bi-level optimization problem, where the number of labeled instances is minimized to reduce the annotating cost, and the target loss is minimized to improve the overall performance.
The practical solution to the bi-level optimization problem presents significant challenges when attempting to solve it directly.
Consequently, it is often necessary to re-formulate this problem as an iterative selection process.

Given $K$ different data sources (domains) $\mathcal{D} = \{\mathcal{D}_1, $ $ \mathcal{D}_2, \dots, \mathcal{D}_K\}$, a set of data pools could be collected $\mathcal{P} = \{\mathcal{P}_1, \mathcal{P}_2,\dots, \mathcal{P}_K\}$. 
The initial labeled and unlabeled data set can be written as $\mathcal{L}_0 = \{\mathcal{L}_{0,1}, $ $ \mathcal{L}_{0,2}, \dots, \mathcal{L}_{0,K}\}$ and $\mathcal{U}_0 = \{\mathcal{U}_{0,1}, $ $ \mathcal{U}_{0,2}, \dots, \mathcal{U}_{0,K}\}$, where $\mathcal{P}_k = \mathcal{L}_{0,k} \bigcup \mathcal{U}_{0,k}$, and $\vert \mathcal{L}_{0,k} \vert \ll \vert \mathcal{U}_{0,k} \vert$ for a domain $k$.
MDAL is to reduce the labeling cost by iteratively selecting informative instances from the unlabeled data set $\mathcal{U}_0$ according to an AL acquisition strategy $\alpha$.
First, a multi-domain model $\mathcal{M}_0$ is trained on the initial labeled data $\mathcal{L}_0$ and unlabeled data $\mathcal{U}_0$.
Then, a batch of to-be-queried instances $\mathcal{Q}_i$ with top-$b$ acquisition utility is selected and annotated by an oracle in the $i$-th AL iteration:

\vskip-10pt

\begin{equation}
  \mathcal{Q}_i = \arg \max_{x \in \mathcal{U}_{i-1}}^{b} \alpha (x, \mathcal{M}_{i-1}), \quad |\mathcal{Q}_i| = b
\end{equation}

\noindent
$\mathcal{L}_{i-1}$ and $\mathcal{U}_{i-1}$ are then updated with the annotated selected batch $\mathcal{Q}_i$.
The model $\mathcal{M}_{i}$ is trained on the updated data as follows:

\vskip-5pt

\begin{equation}
  \begin{aligned}
    \mathcal{M}_i(\mathcal{Q}) & = \min_{\mathcal{M}} {Loss}_{\rm sup}(\mathcal{M}; \mathcal{L}_{i-1} \cup \mathcal{Q}_i) + \Omega (\mathcal{M}; \mathcal{P})
  \end{aligned}
\end{equation}

\noindent
${Loss}_{\rm sup}$ denotes the supervised loss on the labeled data.
$\Omega(\mathcal{M}; \mathcal{P})$ denotes a designed loss on the whole set of data pools $\mathcal{P}$ for capturing the common knowledge through $\mathcal{M}$.
The labeling process terminates once the labeling budget $\mathcal{B}$ is exhausted or the desired performance has been reached.
Finally, the labeled set $\mathcal{L}_{i}$ and the model $\mathcal{M}_{i}$ at the final iteration are obtained as the outputs.

\section{Related Work}

\subsection{Multi-Domain Learning}

Multi-domain learning \cite{MDL} primarily focuses on performing a unified task across multiple domains simultaneously.
Existing MDL research centers around information sharing among domains while preserving domain-specific information, which could be achieved through model architectures.
The widely used architecture for addressing this challenge is the shared-private structure, originally utilized in domain adaptation problems \cite{DSN}.
Adversarial shared-private model (ASP-MTL) \cite{ASP-MTL} is the pioneering approach to employ the shared-private structure in the context of MDL, resulting in significant improvements over single-domain models.
ASP-MTL adopts adversarial learning to encourage domain-invariance in the shared feature extractor and domain-specificity in the private feature extractors.
Several subsequent works \cite{MAN,CAN} have further enhanced the performance based on this share-private architecture.

\subsection{Multi-Domain Active Learning}

Only a limited number of studies directly relate to MDAL.
These works still employ conventional single-domain AL strategies on models trained on multiple domains.
For instance, Li et al. \cite{MDAL-text} first applied active learning with multiple support vector machines on concatenated features from each domain in the context of multi-domain sentiment classification.
He et al. \cite{mdal} conducted a comprehensive comparative study of conventional active learning strategies on multiple neural-network-based MDL models, demonstrating improvements over random selection.
Some works have also applied active learning to multiple domains without considering information-sharing.
They either construct independent classifiers for each domain \cite{mdal-AD} or utilize a single model for all domains \cite{msal}.

In these existing works, information sharing is primarily considered during the model training process, while the selection process evaluates the informativeness of items solely within specific domains.
In other words, conventional AL strategies do not account for the potential impact of a sample on other domains.

\section{Methodology}

\begin{figure}[b]
  \centering
  \vskip-20pt
  \scalebox{0.95}{
    \includegraphics[width=\linewidth]{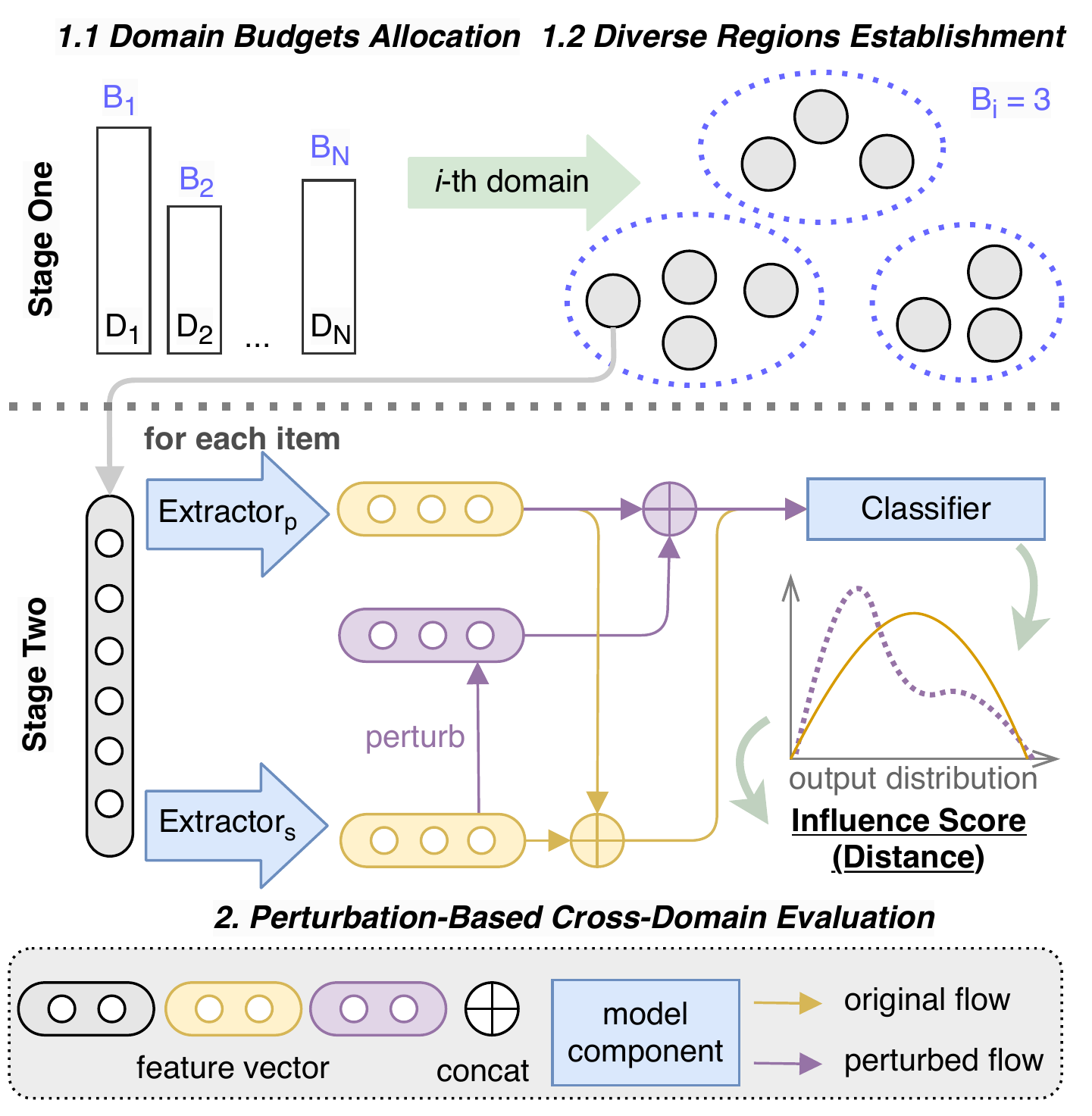}
  }
  % \vskip-5pt
  \caption{An illustration of the proposed P2S-MDAL method. First, selecting regions are established by a budget allocation and a selection space division processes. Then, samples with higher cross-domain influence score would be selected from each region.}
  \label{fig:framework}
\end{figure}

To address the limitations of conventional AL strategies, we propose a novel method P2S-MDAL for MDAL that evaluates the influence of samples on other domains.
P2S-MDAL follows a two-stage framework: selecting regions establishment and perturbation-based item evaluation.
The overall framework is illustrated in Figure \ref{fig:framework}.

\subsection{Selecting Regions Establishment}

To avoid the incomparability between sample evaluation scores from different domains, we constrain the score-based selection within each domain.
Thus, the budgets should be allocated to each domain in advance according to the influence of domains.
% The simplest influence estimate is the number of samples in each domain.
Here we take the total number of samples as an influence estimate.
Let $n_k$ denote the number of samples in domain $k$, and $B$ denote the total budget. 
The budget allocation is calculated as follows:

\vskip-5pt

\begin{equation}
  B_k = \frac{n_k}{\sum_{i=1}^{K} n_i} \times B
\end{equation}

Next, to ensure the diversity of the sample selection, the selection space is divided in each domain with the corresponding allocated budgets.
We employ the $k$-Means algorithm to divide the selection space into $B_k$ regions for the $k$-th domain.
We utilize gradients $E$ at the last layer of the model as embeddings.
Compared to the original feature space, the gradient space is more discriminative and better represents the sample influence on the current model \cite{BADGE}.
The division process could be written as follows:

\vskip-10pt

\begin{equation}
  \{S_{k,1}, \cdots, S_{k,B_k}\} = kMeans(\{E_{k,1}, E_{k,2}, \cdots, E_{k,n_k}\},B_k)
\end{equation}

\noindent
where $E_{k,i}$ represents the gradient of the $i$-th sample in domain $k$, and $S_{k,j}$ represents the $j$-th cluster in domain $k$.

\subsection{Domain Influence Estimation}

Given the established regions for selection, we can evaluate the cross-domain influence of samples in each region, ensuring that the selected samples benefit not only the current domain but also other domains.
The sample with the highest evaluation score is selected from each region.
This evaluation is based on the characteristic of the ASP-MTL model, where a domain-shared feature extractor $F_{s}(\cdot)$ and domain-private feature extractors $F_{p_k}(\cdot)$ are combined to form the final feature representation.
Since the shared information is captured solely by the shared feature extractor, the cross-domain influence evaluation could base on this component.

The intuition of our method is that if a sample is informative to the shared feature extractor, it will be informative to all domains.
In AL, a common approach to evaluating the informativeness of a sample is uncertainty measurement.
Motivated by previous works \cite{DeepFool,feature-squeezing,Saliency-Attack} that measure uncertainty by adversarial samples, we introduce a perturbation-based method to estimate the informativeness of each sample on the shared feature extractor.
If a sample is more informative to the shared feature extractor, it will be more vulnerable to perturbations, i.e., the perturbed sample will be more likely to be misclassified.
Consequently, such examples, which are less learned by the shared extractor, could be more influential to all the domains.
% The rationale is that a more informative instance for all domains should lead to more improvements in the quality of the shared feature extractor.
% should be less learned by the shared extractor, accompanied less robustness to perturbations.
We sample perturbations from a Gaussian distribution $\delta \sim \mathcal{N}(0, \sigma^2)$, and add them to the output of the shared feature extractor.
The perturbed output probability of an item in the $k$-th domain is denoted as:

\vskip-10pt

\begin{equation} \label{eq1}
  \begin{split}
    Out_k(x, \delta) &=  C_{k}((F_s(x) + \delta) \oplus F_{p_k}(x))
  \end{split}
\end{equation}

\noindent
Here, $\oplus$ denotes the concatenation operation, and $C_{k}$ represents the classifier for the $k$-th domain.
The distance between the original output and the perturbed output is used to evaluate the cross-domain informativeness of the sample.
The distance is calculated by the Kullback-Leibler divergence \cite{KL} between two distributions, which can be written as:

\vskip-5pt

\begin{equation} \label{eq2}
  \begin{split}
    % Score(x,{\theta_s}) &= \underset{\delta}{\mathbb{E}}\left[Distance(Out_i(x), Out_i(x, \delta))\right],\\
    Score(x) &= \underset{\delta}{\mathbb{E}}\left[Distance(Out_i(x), Out_i(x, \delta))\right],\\
    Distance(P,Q) &= D_{\mathrm{KL}}(P \| Q) = -\sum_{x \in \mathcal{X}} P(x) \log \left(\frac{Q(x)}{P(x)}\right)
  \end{split}
\end{equation}

\noindent
% where $\theta_s$ represents the parameters of the shared feature extractor.
The score represents the expected output distance between the original and perturbed output distributions.
Empirically, the score is calculated by sampling multiple perturbations.

\section{Experiments}

\subsection{Research Questions}

\begin{enumerate}
  \item Whether P2S-MDAL improves the performance of the ASP-MTL model compared to conventional strategies?
  \item Whether each stage of P2S-MDAL provides positive effects?
  \item Whether P2S-MDAL is applicable in terms of time?
\end{enumerate}

\subsection{Experimental Setup}

\begin{figure*}[htbp]
  \centering
  \scalebox{0.94}{
  \hspace{-0.03\linewidth}
  \subfigure[Amazon]{
    \begin{minipage}[b]{0.27\linewidth}
      \includegraphics[width=1\textwidth]{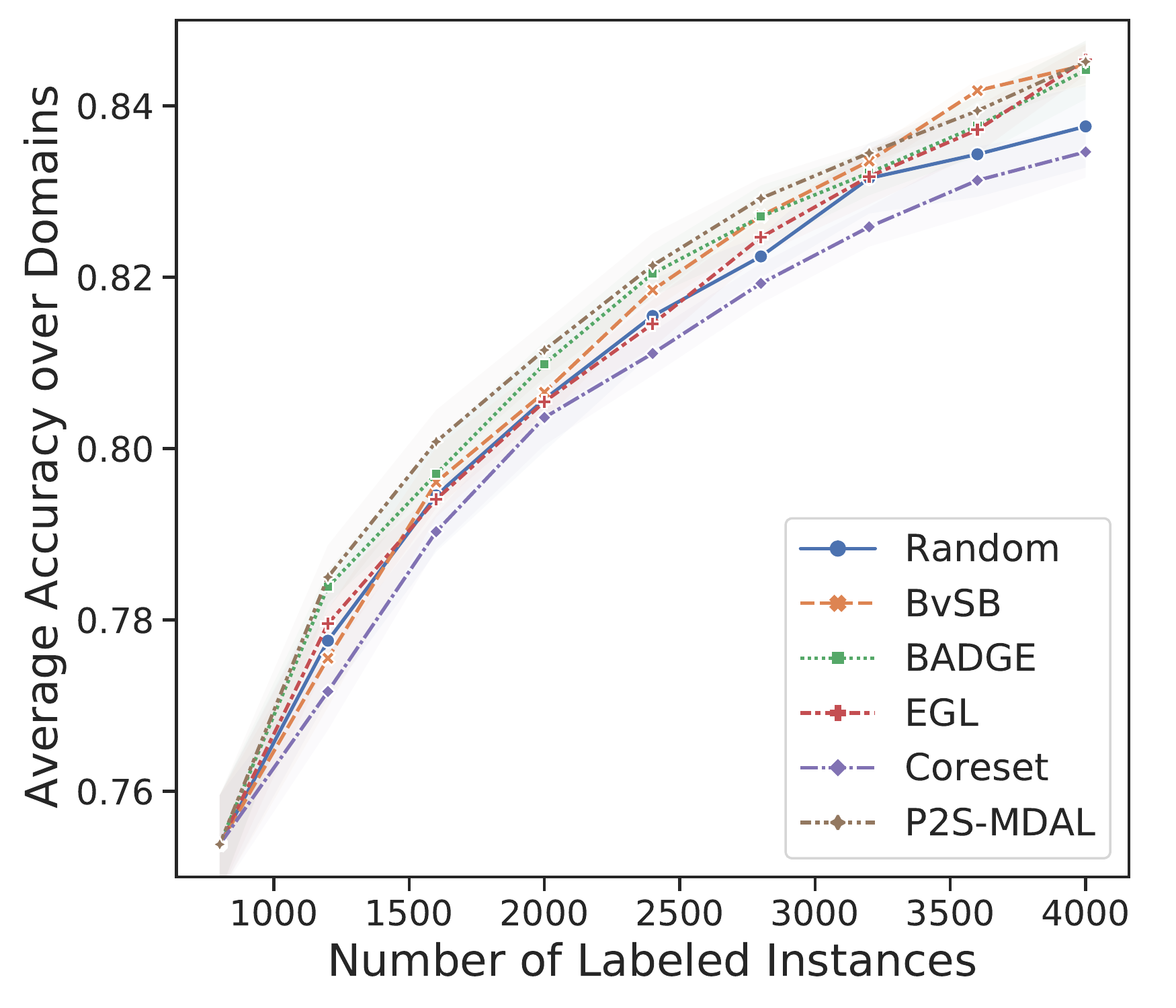}
    \end{minipage}
    \label{fig:main-results-amazon}
  }
  \hspace{-0.025\linewidth}
  \subfigure[COIL]{
    \begin{minipage}[b]{0.27\linewidth}
      \includegraphics[width=1\textwidth]{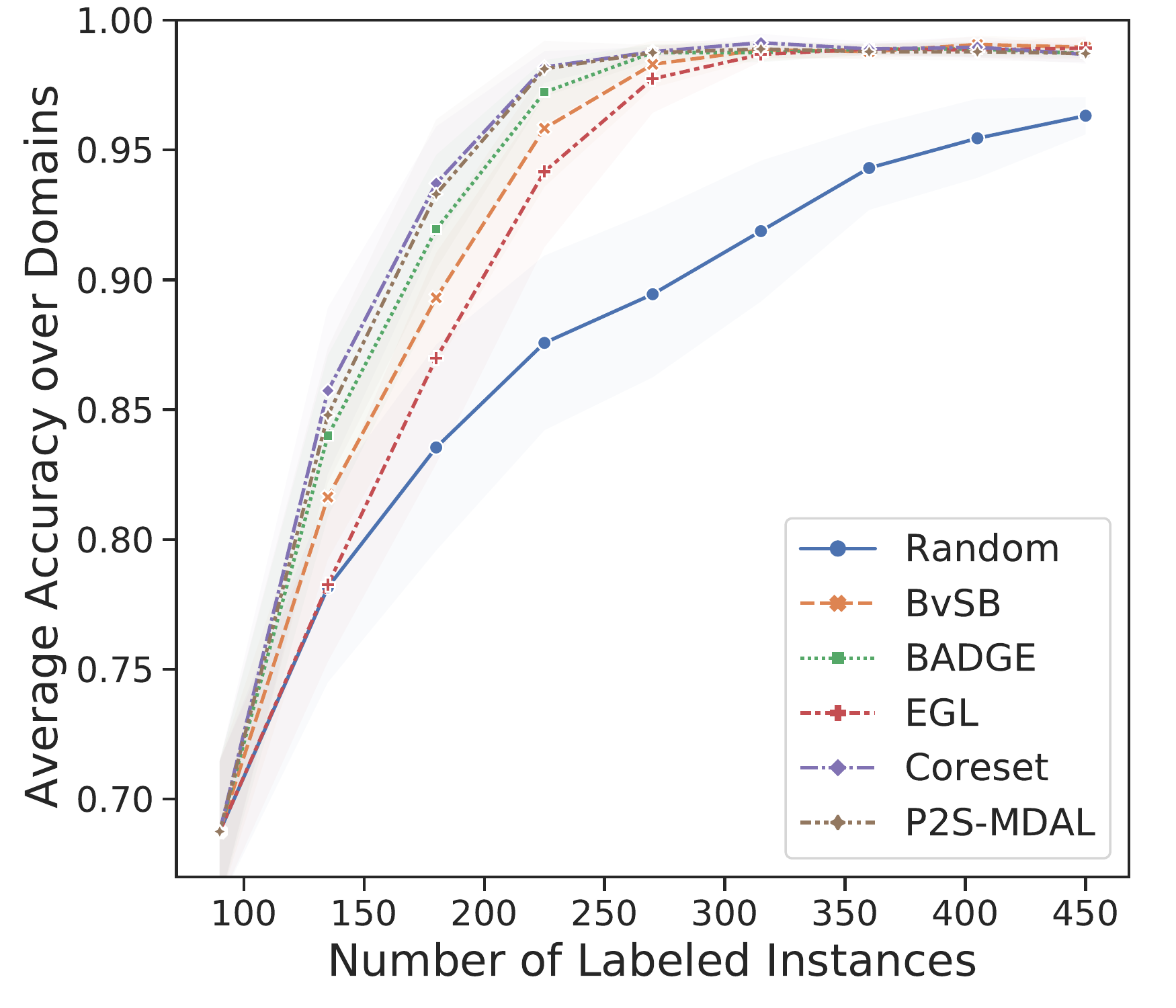}
    \end{minipage}
    \label{fig:main-results-coil}
  }
  \hspace{-0.025\linewidth}
  \subfigure[FDUMTL]{
    \begin{minipage}[b]{0.27\linewidth}
      \includegraphics[width=1\textwidth]{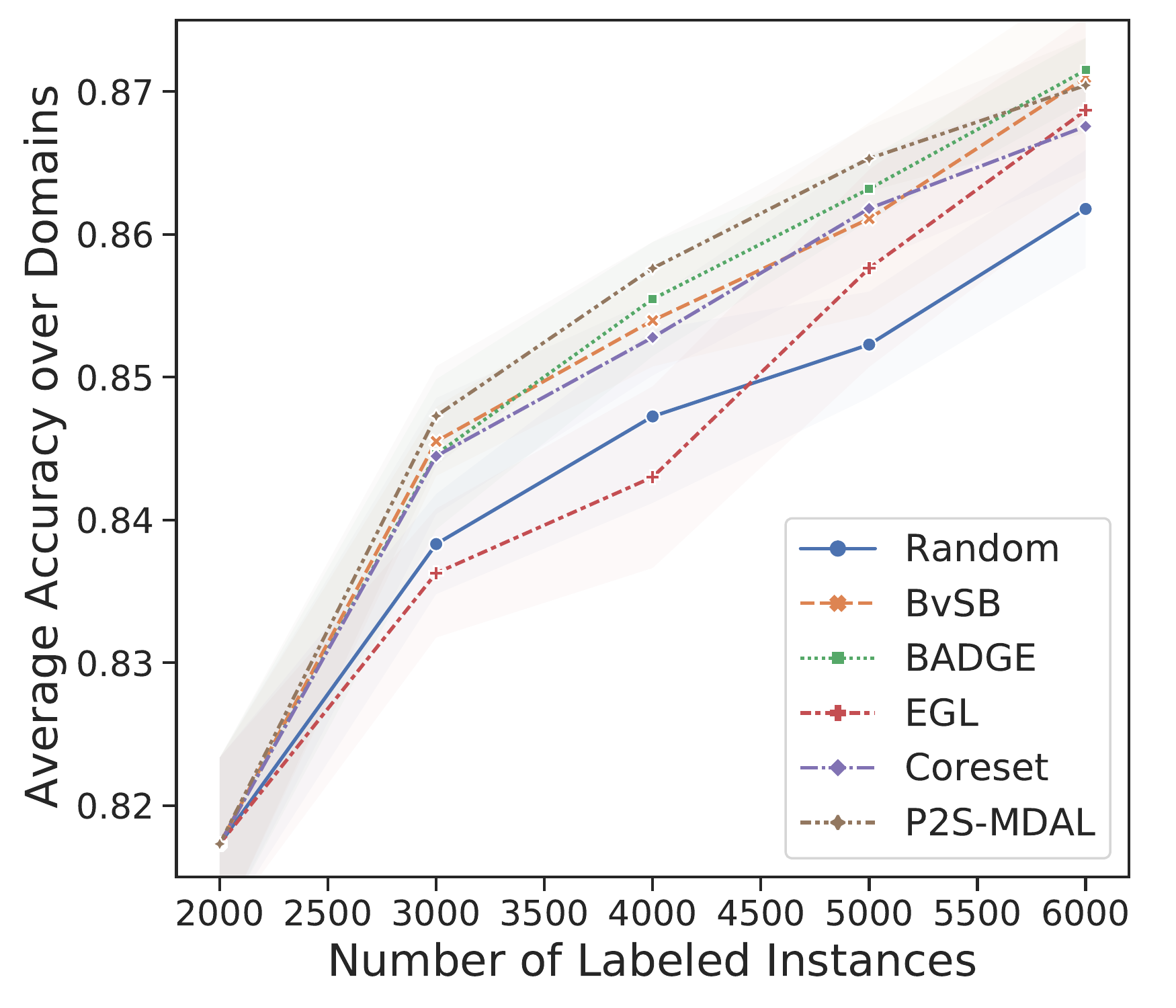}
    \end{minipage}
    \label{fig:main-results-fdumtl}
  }
  \hspace{-0.025\linewidth}
  \subfigure[Ablation Study on FDUMTL]{
    \begin{minipage}[b]{0.27\linewidth}
      \includegraphics[width=1\textwidth]{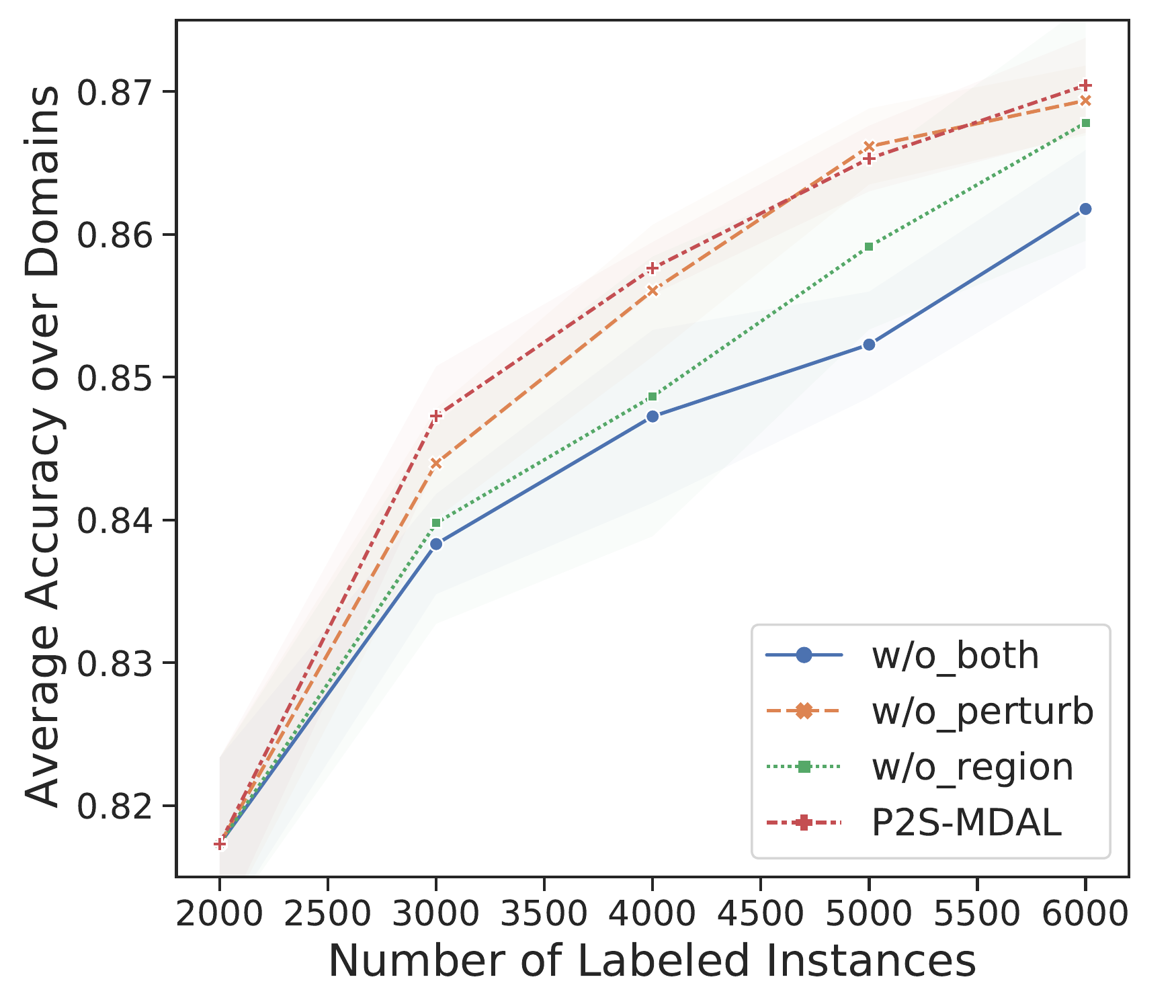}
    \end{minipage}
    \label{fig:ablation-study}
  }
  }
  \vskip-10pt
  \caption{Performance in terms of learning curves. (a)-(c) are the results on three datasets. (d) is the ablation study on FDUMTL.}
  \vskip-8pt
  \label{fig:main-results}
\end{figure*}

\subsubsection{Dataset}

Three popular multi-domain textual and image datasets are used in our experiments, namely Amazon \cite{mSDA}, COIL \cite{COIL}, and FDUMTL \cite{ASP-MTL}.
Amazon dataset consists of four textual domains, each containing two categories, with instances encoded to a vector representation of length 5000.
COIL dataset consists of two image domains, each containing twenty categories, with instances encoded to a vector representation of length 1024.
FDUMTL dataset comprises sixteen textual domains, each containing two categories, with raw texts utilizing word2vec embedding.

\subsubsection{Model Implementation}

ASP-MTL \cite{ASP-MTL} is used with the proposed strategy.
On datasets COIL and Amazon, single hidden layer MLP with width 64 is used.
On dataset FDUMTL, CNN is used as feature extractor with the output dimension 128.
The classifier is a fully connected layer for all datasets.
The model is trained using an SGD optimizer with batch size 8.

\subsubsection{AL settings \& Evaluation}

Five conventional single domain AL strategies are selected for the comparison.
\textbf{Random} is the simplest strategy, which randomly selects instances from each domain.
\textbf{Best vs. Second Best (BvSB)} \cite{BvSB}, as an uncertainty measurement, selects instances with the greatest difference in prediction probability between the most and second most likely classes.
\textbf{Expected Gradient Length (EGL)} \cite{EGL,EGL_text} is designed for models that can be optimized by gradients.
The instances leading to the longest expected gradient length to the last fully connected layer will be selected.
\textbf{Coreset} \cite{coreset} selects instances using a greedy furthest-first search conditioned on the currently labeled examples by using the middle representation.
\textbf{Batch Active learning by Diverse Gradient Embeddings (BADGE)} \cite{BADGE} calculates the gradients of the last fully connected layer.
A $k$-Means++ initialization is applied to the gradients to ensure the diversity of the selected batch.

In the selecting process, for Amazon and COIL dataset, 10\% of the training set are annotated to initialize the model.
The total budget is 50\% of the training set, and 5\% of the training set is selected in each iteration.
For FDUMTL, the total budget is 30\% of the training set.
For the proposed AL strategy, the perturbation is sampled 20 times with amplitude of 0.01.
All the experiments are repeated 5 times with different random seeds to get an average performance.
The learning curves are used to evaluate the performance of the AL strategies, where the $x$-axis represents the number of selected instances, and the $y$-axis represents the accuracy of the model on the test set.
The area under learning curve (AULC) is also calculated.

\subsection{RQ1: Performance Evaluation}

A comparative analysis was undertaken between the proposed approach and five AL strategies, utilizing three datasets.
The results are shown in Figure \ref{fig:main-results-amazon}-\ref{fig:main-results-fdumtl}.
Notably, the performance of a single strategy can vary greatly across different datasets.
For instance, the proposed method and Coreset share the top tier in performance on the COIL dataset.
However, Coreset's performances on Amazon and FDUMTL datasets are suboptimal.
In contrast, P2S-MDAL consistently performs at an optimal level across all three datasets.
Besides, for FDUMTL dataset, with its sixteen domains, the selection hardness grows in complexity.
P2S-MDAL still outperforms others on this dataset, thereby further demonstrating its effectiveness.

\begin{table}
  \centering
  \caption{Performance in terms of AULC of five conventional AL strategies on three datasets.}
  \vskip-10pt
  \label{tab:performance-AULC}
  \scalebox{0.9}{
    \begin{tabular}{c|c|c|c}
      \toprule
      Strategy/Datasets    & Amazon            & COIL              & FDUMTL            \\
      \midrule
      Random               & 80.97(0.41)       & 87.66(1.78)       & 84.43(0.33)       \\
      BvSB                 & 81.23(0.09)       & 92.88(0.9)        & 85.12(0.26)       \\
      BADGE                & 81.34(0.16)       & 93.92(0.76)       & 85.19(0.22)       \\
      EGL                  & 81.09(0.14)       & 91.93(1.33)       & 84.5(0.45)        \\
      Coreset              & 80.59(0.16)       & \pmb{94.52(0.49)} & 85.04(0.26)       \\
      \pmb{P2S-MDAL(ours)} & \pmb{81.52(0.13)} & 94.48(0.69)       & \pmb{85.35(0.18)} \\
      \bottomrule
    \end{tabular}
  }
  \vskip-10pt
\end{table}

\subsection{RQ2: Ablation Study}

First, the effectiveness of the proposed perturbation module is analyzed.
Utilizing the same selecting regions establishment processes (first stage), various AL strategies (namely Center, BvSB, EGL) are substituted at the second stage.
The performance, measured in terms of AULC, is depicted in Table \ref{tab:perturbation-analysis}.
Once again, P2S-MDAL comes out as the top performer, underscoring the efficacy of the perturbation module.

\begin{table}
  \centering
  \caption{Perturbation Analysis on FDUMTL dataset.}
  \vskip-10pt
  \label{tab:perturbation-analysis}
  \scalebox{0.9}{
    \begin{tabular}{c|c}
      \toprule
      Two-Stage Strategy   & AULC               \\
      \midrule
      2S-Center            & 85.14 (0.16)       \\
      2S-BvSB              & 85.12 (0.24)       \\
      2S-EGL               & 84.98 (0.21)       \\
      \pmb{P2S-MDAL(ours)} & \pmb{85.35 (0.18)} \\
      \bottomrule
    \end{tabular}
  }
  \vskip-1pt
\end{table}

An ablation study was also carried out to further analyze the effectiveness of both stages.
Specifically, we removed the perturbation module (w/o perturb) and the region establishment module (w/o region) individually, and subsequently compared their performances.
The results of this study are illustrated in Figure \ref{fig:ablation-study}.
With both components, P2S-MDAL achieves the best performance.

% \begin{figure}[htbp]
%   \centering
%   \scalebox{0.75}{
%     \includegraphics[width=\linewidth]{images/fdumtl-ablation.pdf}
%   }
%   \caption{Ablation Study.}
%   \label{fig:ablation-study}
% \end{figure}

\subsection{RQ3: Time Complexity Analysis}

The time complexity is qualitatively analyzed in this section.
% P2S-MDAL is a two-stage approach, which can be considered a combination of conventional score-based method or distribution-based method.
The time complexity for score-based methods (e.g. Uncertainty, EGL) typically falls in $O(n)$, where $n$ represents the quantity of unlabeled instances.
P2S-MDAL includes the scoring stage, which means it at least has higher time complexity than score-based methods.
Distribution-based methods (e.g. BADGE, Coreset) generally have a high time complexity due to the requisite calculations of pairwise distances.
% Consequently, these methods become more time-consuming as the number of domains and instances increase.
P2S-MDAL also calculates pairwise distances in selecting regions establishment, whereas it is implemented on each domain with a reduced number of items.
This leads to significantly shorter processing times when compared to distribution-based methods.

\section{Conclusions and Future Directions}

% In this paper, we introduce an innovative P2S-MDAL strategy to reduce labeling burden, which is tailored for multi-domain data.
This paper marks the first instance of a dedicated AL strategy designed to address the MDAL problem.
The proposed P2S-MDAL method is a two-stage approach, which first establishes selection regions in each domain for sampling diversity, and then selects the most cross-domain influential instances in each region by using perturbations.
% It begins with budget allocation and data division into various clusters, followed by the selection of the most influential cross-domain instances within each cluster.
The efficacy of the proposed method is substantiated by the evaluations on three separate datasets.
Furthermore, the ablation study and the perturbation analysis contribute additional proof of the effectiveness of each module in P2S-MDAL.

Looking toward the future, we anticipate that the budget allocation method could be enhanced by incorporating considerations of data distribution and domain difficulty.
With the feedback from either the training or validation set, the budget allocation could be adjusted to better fit the data.
Moreover, with some adaptations, the proposed perturbation-based cross-domain evaluation could be extended to other MDL models.
Furthermore, we envision the development of a unified MDAL method, where both models and strategies are designed within single framework.
This requires the model training to incorporate the unique characteristics of the AL-selected items.
Meanwhile, the AL selection process could benefit from the explicitly designed structure of the model.


\newpage

\begin{acks}
  % N/A.
\end{acks}

\bibliographystyle{ACM-Reference-Format}
\bibliography{mdal-strategy}

\end{document}